# Shape-Based Plagiarism Detection for Flowchart Figures in Texts

Senosy Arrish, Fadhil Noer Afif, Ahmadu Maidorawa and Naomie Salim

Faculty of Computing, Universiti Teknologi Malaysia, Skudai, Malaysia

**ABSTRACT**

*Plagiarism detection is well known phenomenon in the academic arena. Copying other people is considered as serious offence that needs to be checked. There are many plagiarism detection systems such as turn-it-in that has been developed to provide this checks. Most, if not all, discard the figures and charts before checking for plagiarism. Discarding the figures and charts results in look holes that people can take advantage. That means people can plagiarized figures and charts easily without the current plagiarism systems detecting it. There are very few papers which talks about flowcharts plagiarism detection. Therefore, there is a need to develop a system that will detect plagiarism in figures and charts. This paper presents a method for detecting flow chart figure plagiarism based on shape-based image processing and multimedia retrieval. The method managed to retrieve flowcharts with ranked similarity according to different matching sets.*

## 1. INTRODUCTION

Flowcharts become a significant issue to explain different kinds of information based on figure types. In some documents, flowcharts are so important to illustrate a lot of details and make it easier to understand. Some of complex problems cannot be solved directly without explain these problems in flowcharts. The most significant role of using flowcharts is in the designing part of projects. In many projects, methodology of structured design is one of primary steps to build entire system and solving engineering problems that can be explained by using flowcharts and other types of figures. In addition to previous points, using flowchart in design helps to divide problem into smaller parts and manage problems easily.

In the last few years, Plagiarism detection became so essential topic in researches area. Plagiarism can be involve in different fields research papers, art area and program code. There are several definitions of plagiarism, for example: taking someone else work then makes it as a work of your own. This process consider as plagiarism. The growth of plagiarism become very common because of the number of available resources increased. Resources such as internet and journals make plagiarize easy for users [1].

There are expansion of online plagiarism detection services because of the increased of research plagiarism detection needs of education research and publication. Using internet services to detect on plagiarism make it easier than using other detection services [2]. There are many researches discussed plagiarism detection of document like text, figure, image, so that several techniques and methods are proposed to detect about plagiarism in these types of documents. However, while text-based plagiarism detection systems are now become common, plagiarism detection for flowcharts have not been explored much by researchers.

Plagiarism detection is well known phenomenon in the academic arena. Copying other people is considered as serious offence that needs to be checked. There are many plagiarism detection systems that has been developed to provide this checks. Most, if not all, discard the figures and charts before checking for plagiarism. Discarding the figures and charts results in loop holes that people can take advantage from. That means people can plagiarize figures and charts easily without worrying to be detected by the current text-based plagiarism systems. Therefore there is a need to develop a system that will detect plagiarism in figures, particularly flowcharts.

## 2. RELATED WORKS

Currently, there are only a few researches working on the issue of figure plagiarism, particularly diagrams and flowcharts. However, despite not many works have been presented in flowchart plagiarism system, there are some works related to this issue, such as methods to characterize flowchart types based on the image features. Miyao and Maruyama used loop structure as source to detect directional strokes into learning system [3]. The method managed to recognize 97.6 % of the given data sets, however it is still not fully practical as the system requires the user to draw a loop structure by himself.

A flowchart conveys information about process of works to the reader. Consequently, the text contained in the charts may be used as features of description. Vasudevan et al. presented a method to extract information inside flowcharts by contour processing and neural network-based optical character recognition [4]. The method analyzes flow lines and contour of the system as feature descriptors. The system managed to provide recall rate of 98%. Awal et al. took a different approach to characterize flowcharts semantically using grammatical approach [5].
Aside from information contained in a flowchart, the shapes used by one node of flowchart is of equal importance, as it can distinguish between different processes in the work. It can be stated that flowchart is a subset of images, thus it can be characterized by kinds of image features. Furthermore, to develop a flowchart plagiarism system, the system should be able recognize these features which will answer required query by human. Zhang and Lu reviewed techniques to represent and describe figures based on their shape features [6]. The method depends on the feature inside each document like color, shape, and texture. In addition to color and textures, there are number of common representation of features vector, strings, and graphs, fuzzy and probabilistic representation that can be used to describe a figure [7].

## 3. METHODOLOGY

The main goal of this project is to create a figure plagiarism system that is primarily based on shape. This system primarily focuses on flowcharts detection. The database contains flowchart images stored in a single folder. The system will retrieve and rank this database based on a given query by the user. The retrieval system works by detecting shapes in each figure and compare to the shape from the query, as shown in Figure 1.

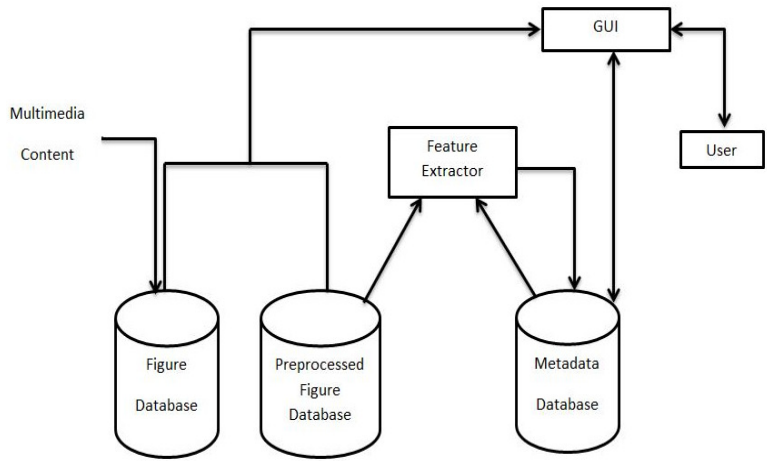

Figure 1.  Multimedia Information Retrieval System.

### 3.1. Pre-processing

In order for the system to obtain maximum retrieval result, the obtained figures need to be pre-processed. The pre-processing is done to reduce retrieval errors and help the system accuracy. There are three steps on pre-processing will be done:

1.   Thinning
2.   Removing connected lines
3.   Removing text

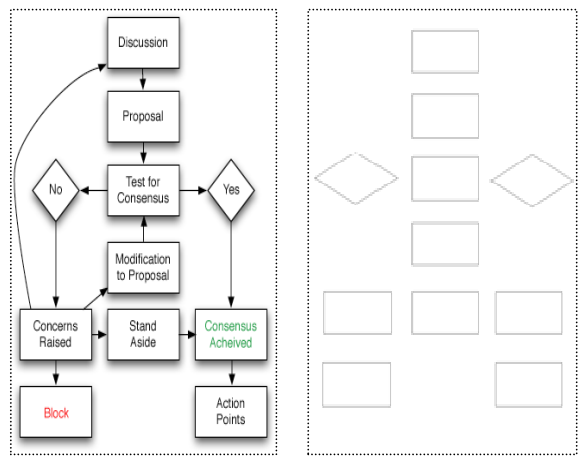

Figure 2 Processing Example

### 3.2.  Database

To create the database, first is to group figures and preprocessing figure into separated sets.  The databases are created by two sets of databases:

115

International Journal of Computer Science & Information Technology (IJCSIT) Vol 6, No 1, February 2014

1. First database is for storing the figures.
2. Second database is for storing the preprocess figure.

Each database consists of figures and preprocessed figure. Each of the figures has been given an id. This id is use for information retrieval and also to be recall back to the metadata. These projects only consider four basic shapes in flowcharts which are connector, process, decision, and start/stop. **Figure 3** gives an example of the figure from the database and example of the figure from the database.

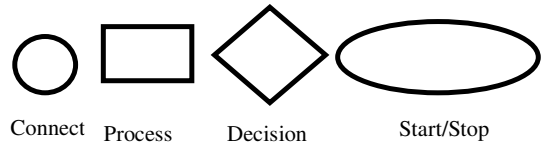

Connect   Process   Decision   Start/Stop

Figure 3. Sample Database Shapes

### 3.3. Metadata Extraction

From the database the shape can trace back to the metadata. The process of this step is getting the image queries and extracts index shape inside query figures. The same as in database for the metadata each figure has been given an id. The figures can now be stored to the database as well as the metadata, as shown in Table 1.

Table 1. Sample Metadata Database

| FigureID | Connector | StartStop | Decision | Process |
|---|---|---|---|---|
| 1 | | | | |
| 2 | | | | |
| 3 | | | | |
| 4 | | | | |

The use of the table is to speed up the process of the detection of figure and the comparison without consulting the figures in the database. The metadata provide some structure that improves the analyzing speed using the figure's low level features. Each metadata entries is number of particular shapes obtain by shape detection technique.

The shape detection for this project uses area detection technique. This measures distance from the center point of the shape to its boundaries. This distance will be compared with mathematical formulation of basic shape to retrieve the corresponding shape. Algorithms for this technique is the following:

1. Load the preprocess figure
2. Convert to gray scale
3. Detect edges using canny edge detection
4. Localize the each shape in the figure
5. For each localized shape
    a. Determine the centroid.
    b. Determine the boundaries
    c. Calculate the distance from centroid to boundary using Euclidean distance.
    d. Let A = highest distance
       Let B = lowest distance
       Let C = number of pixel in each shape
    e. Circle = A – B
    f. Ellipse = C / A * B * π
    g. Rectangle = C / ( 4 * B ( $A^2 – B^2$ ) $^{0.5}$)
    h. Diamond = ( C * ( $A^2 – B^2$ ) $^{0.5}$) / 2 * $A^2 B$
    i. If { ( circle < 10 )
          Shape = circle
          Else if ( ellipse < 1.05) and ( ellipse > 0.95)
       `  Shape = ellipse
          Else if ( rectangle < 1.05) and ( rectangle > 0.95)
          Shape = rectangle
          Else if ( diamond < 1.05) and ( diamond > 0.95)
          Shape = diamond
                                                                }

## 3.4. Plagiarism Retrieval Search Engine

The core of the figure plagiarism detection system is the plagiarism retrieval search engine. The search engines receive a user query in the form of sample figure. It takes the sample figure and pre-processes it to build the query vector that will be compared with the figure-document vectors on the metadata database. Similarity between the query vector and each figure-document is calculated using the cosine similarity measure. The most similar figures are retrieved from the database using their figure id, ranked and presented to the user. Figure 4 below illustrates the setup.

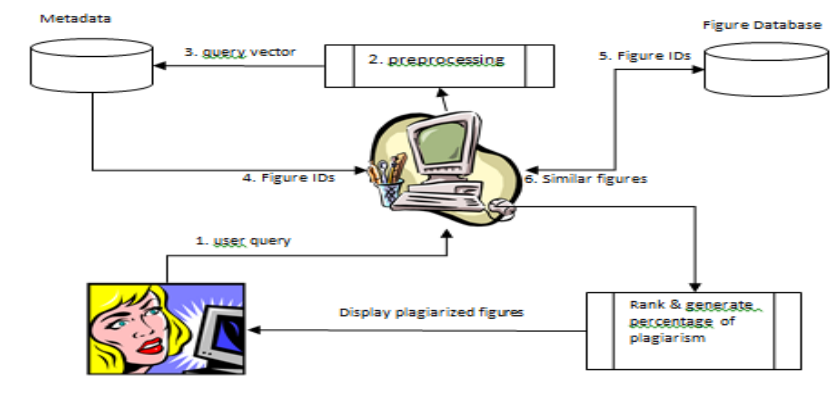

Figure 4 . Search Engine Flowchart

## 3.5. User Query

The user query is in the form of sample figures which can be supplied to the system individually or be extracted from a specified document. Low level features of the query figure are extracted similar to the database figures as highlighted in the **Figure 5** below.

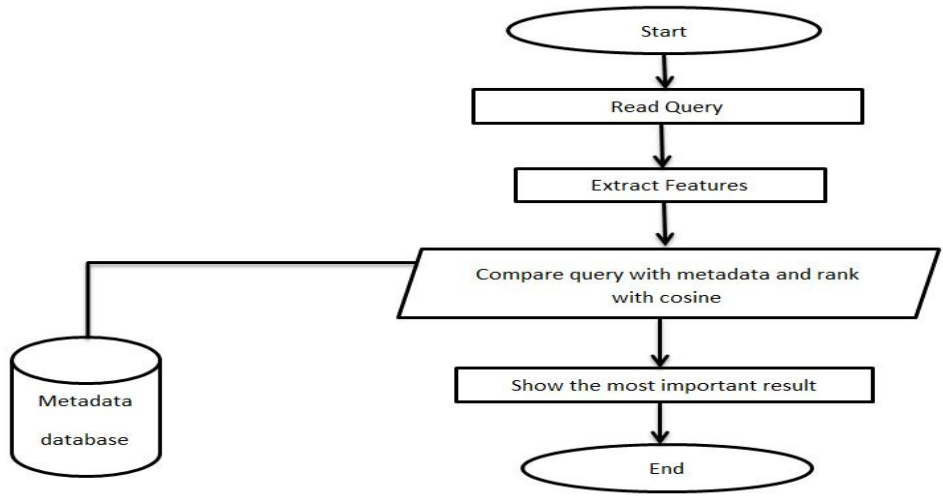

Figure 5 Query Feature Extraction Steps

This system is using Query-By-Example. Figure 6 shows a typical flowchart figure that can be supplied to the plagiarism detection system as a query.

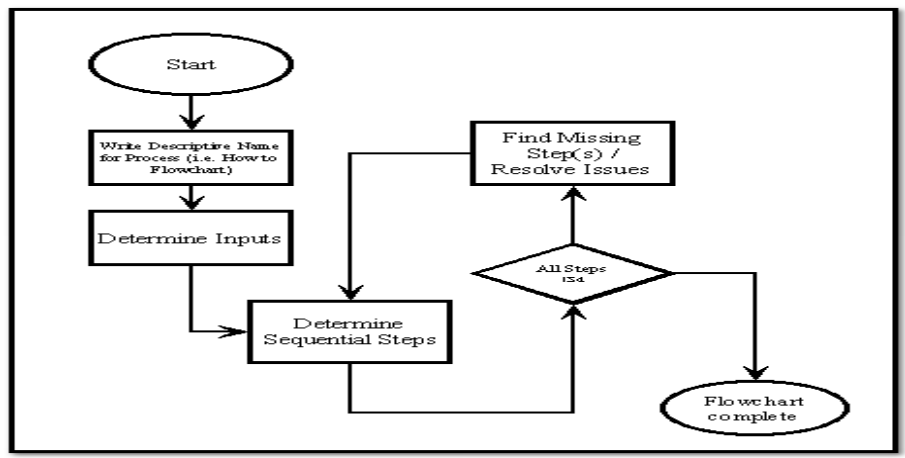

Figure 6. A typical sample query flowchart

## 3.6. Preprocessing of Query

The essence of this stage is to build the query vector that is based on index shape. It starts by removing the stop shapes, that is arrows which are common to all flowcharts and discarding the text inside the various shapes. Shapes are identified using the chain-code technique. Finally the query vector is built. Figure 7a and Figure 7b displayed the outcome of removing the stops shape and the generated query vector respectively.

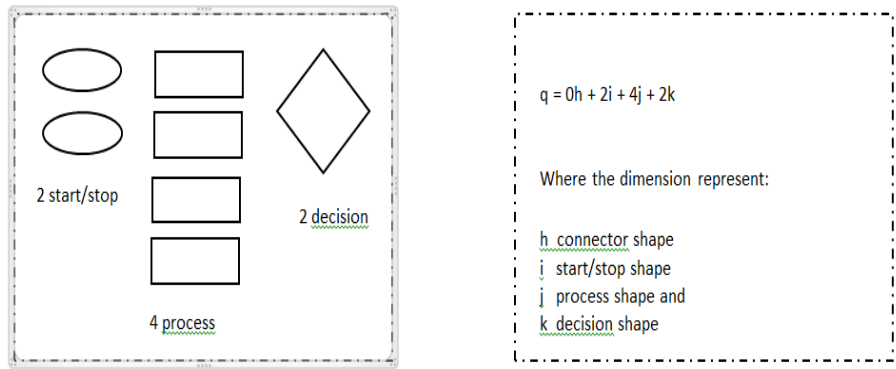

Figure 7. Stop removal and shape detection.

## 3.7. Query and Figure Documents Comparisons

Figure documents descriptions are stored on the metadata database. Each figure document vector has the same dimension with the query vector. Since both the figure documents vectors and the query vector have the same dimension, they can be compared with the query vector one by one using the cosine similarity measure as follows:

$$sim(q, d_j) = \cos(\theta) = \frac{vec(d_j) \cdot vec(q)}{|vec(d_j)| \times |vec(q)|}$$

Where θ is the angle between the figure document vector and the query vector. The cosine values lie between 0 to 1, that is

0 <= cos (θ)<= 1

Where 0 indicates that no similarity between the two vectors and 1 indicates the two vectors are identical. The similarity values are used to rank the figure documents to be retrieved and also used in calculating the percentage of plagiarism on the query figure or figures. The figure IDs of all figure documents with similarity value above a threshold (say above 0.3) are sent to the search engine which will in turn used the figure IDs to retrieve the plagiarized figures from the figure database. See Figure 8 below:

International Journal of Computer Science & Information Technology (IJCSIT) Vol 6, No 1, February 2014

Figure 8. Retrieval Workflow

## 4. RESULT AND ANALYSIS

### 4.1. Datasets

In this paper, datasets used consists of twenty figures of flowcharts obtained from internet resources, containing various amount of shapes. All entries in the datasets were preprocessed to obtain the shape figure, with example shown in **Figure 9** . The metadata generation step will create low-level features of these figures and store it as a vector space in the metadata. When a query arrives from user, it will be evaluated with the metadata to obtain the similarity rank.

Figure 9. Result of Sample

International Journal of Computer Science & Information Technology (IJCSIT) Vol 6, No 1, February 2014

## 4.2. Retrieval Result and Analysis

In order to validate the system's objective, different types of query have been used. The first type is a query that has exact match with an entry in the database.

### 4.2.1  Exact Matching Query

The first experiment used an exact match of query. Any figure in the database will be used to validate the retrieval. Shown in Figure 10, the flowchart was used as a query and run into the retrieval system.

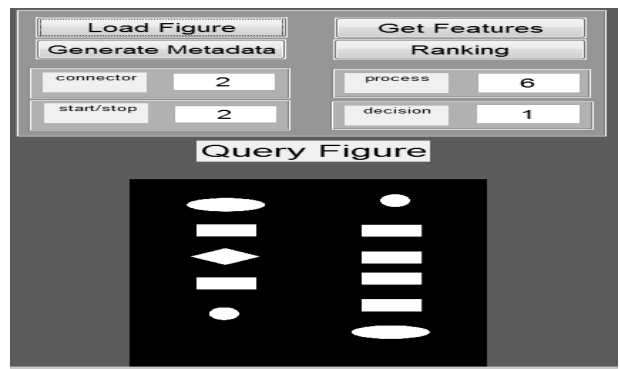

Figure 10. Figure used as query.

The system managed to correctly recognize a total of 11 activities in the figure. The next step is to determine the similarity value with the database and rank them based on it. Four most similar figures were retrieved by the system and displayed as shown in Figure . According to result given by the system, the first rank returns a figure with similarity value 1, then the next rank return 0,94 similarity and so on. Similarity value of 1 in the retrieval shows that the first retrieved figure is the exact match of the query. This was also verified by directly checking to the source database.

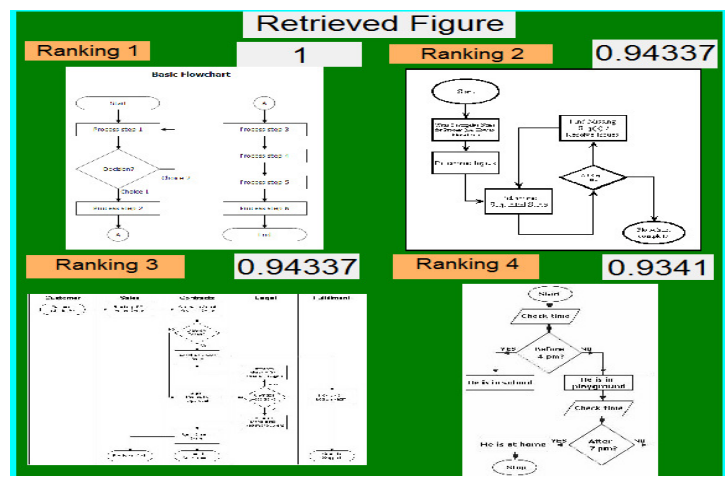

Figure 11. Four most similar figure for a given query.

International Journal of Computer Science & Information Technology (IJCSIT) Vol 6, No 1, February 2014

Figure 4 shows the similarity value for all the retrieved ranks. From the plot it can be seen that the first rank returned the highest similarity value among all figures inside the database, which validates the objective of a retrieval system. Furthermore, similarity value went lower for the next subsequent rank until it reached the lowest value of 0.59, meaning that the last rank has the least similarity with the query.

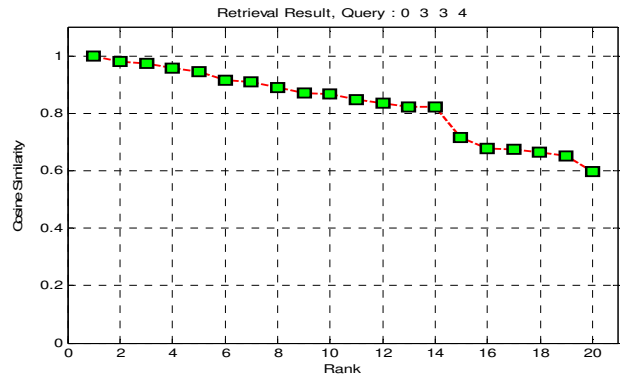

Figure 12. Similarity values of the first Query

### 4.2.2 Partial Matching Query

The second type of query is a partial matching. In this experiment, the query used is a figure that contains partially same shape distribution with one of the database.

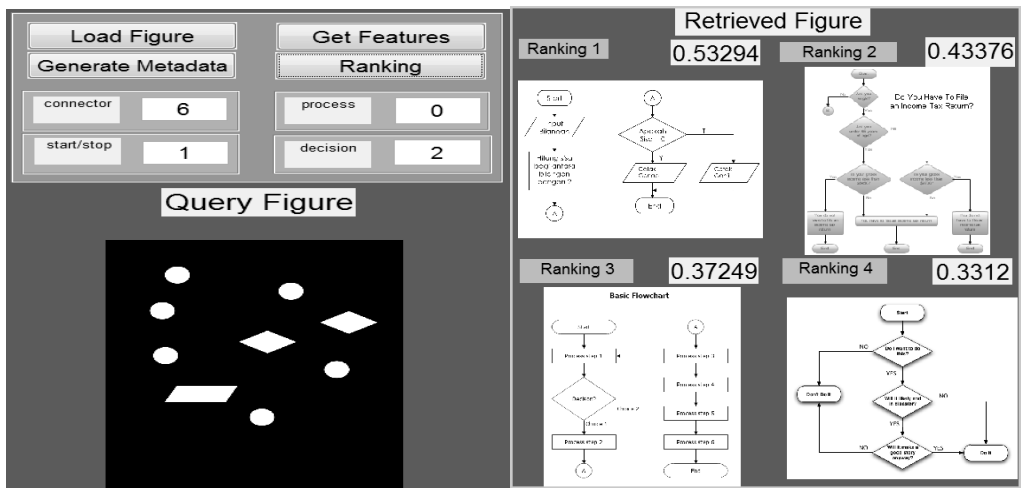

Figure 13. Partial matching query and top retrieved rank.

From the query, a total of 9 activities were recognized by the system. In the result shown by Figure 5, the system gave different result with the first query. It can be seen that the first rank returns 0,532 similarity value. In the graph shown by , the retrieval value ranges from 0,532 from the lowest 0,03. From the result it can be derived that the system returns lower maximum similarity than the first query because the query is only partially matched with the database.

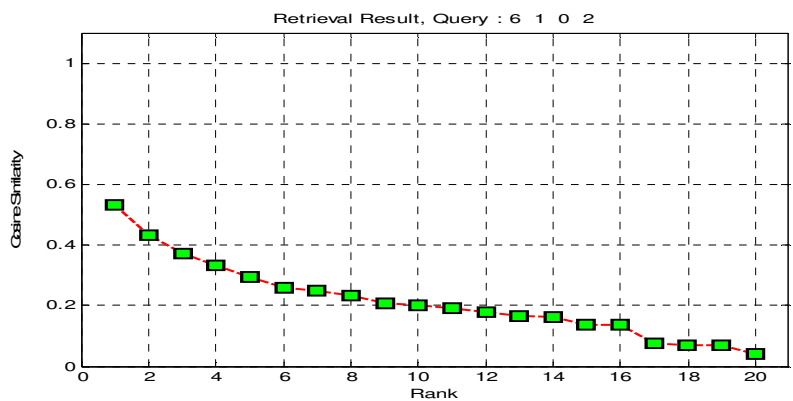

Figure 14. Similarity values of the second query.

## 5. CONCLUSION

In this paper, the system is developed to retrieve figures having certain characteristics based on the given query. The system should be able to retrieve figure with the most similarity value. As was done with the first query, the system managed to return similarity value of 1 with an exact match figure. Moreover, system returns similarity value lower than 1 when a partially match query is given by the user as was done with the second query. When there is nearly unexact match of query in the system, as was indicated in the third query, the system returns similarity value close to zero.

## ACKNOWLEDGEMENTS

This research is supported by Faculty of Computing, Universiti Teknologi Malaysia.

**Authors**

Senosy Suliman Mohamed Arrish
DOB 18/03/87
Bachelor of computer science, Sebha university,Libya, 2009
Master of computer science UTM university,Malaysia, 2014.

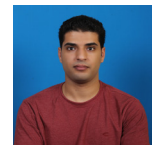

Fadhil Noer Afif
Bachelor of Science, Padjadjaran Un iversity
Master of Science (Computer Science) Universiti Teknologi Malaysia, 2013.

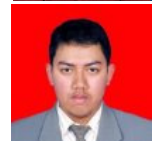

Ahmadu Maidorawa
DOB: 18/ 04/72
QUALIFICATION: B. Tech (2000) Comp sci @ ATBU, MSc (2013) comp sci @ UTM

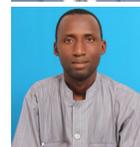

Dr. Naomie Salim is currently a Professor at the Faculty of Computing, Universiti Teknologi Malaysia. She has a Bachelor of Science (Computer Science) from Universiti Teknologi Malaysia. She obtained her Masters Degree in Computer Science from Western Michigan University, USA and her PhD in Information Studies (Chemoinformatics) from University of Sheffield. She taught at both undergraduate and postgraduate level in subjects related to Databases and Information Systems. Her research interest includes Information Retrieval and Cheminformatics. Dr Naomie has been involved in 28 research projects and authored over 150 journal articles and conference papers describing research into novel techniques for computerised information retrieval, with particular reference to textual, chemical and biological information.

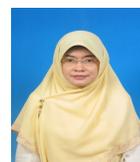